# A Field Guide to Deploying AI Agents in Clinical Practice

## Authors:

Jack Gallifant[1,2], Katherine C. Kellogg[3], Matt Butler[1,4], Amanda Centi[1], Shan Chen,[1,2] Patrick F. Doyle[1,5], Sayon Dutta[1,2], Joyce Guo[1], Matthew J. Hadfield[6], Esther H. Kim[1], David E. Kozono[1,2,5], Hugo JWL Aerts[1,2,7], Adam B. Landman[1], Raymond H. Mak[1,2,5], Rebecca G. Mishuris[1], Tanna L. Nelson[8], Guergana K. Savova[2,10], Elad Sharon[2,5], Benjamin C. Silverman[1,2], Umit Topaloglu[8], Jeremy L. Warner[6,9], Danielle S. Bitterman[1,2,5*]

## Affiliations:

[1] Mass General Brigham, MA, USA
[2] Harvard Medical School, MA, USA
[3] MIT Sloan School of Management, MA, USA
[4] Brown University Health, RI, USA
[5] Dana-Farber Cancer Institute, MA, USA
[6] The Legorreta Cancer Center at Brown University, Brown University Health Cancer Institute, Providence, RI, USA
[7] Maastricht University, The Netherlands
[8] Center for Biomedical Informatics & Information Technology, National Cancer Institute, NIH, MD, USA
[9] Rhode Island Hospital, Providence, RI, USA
[10] Computational Health Informatics Program (CHIP), Boston Children's Hospital, MA, USA

## Corresponding Author:

Dr. Danielle S. Bitterman
Department of Radiation Oncology
Dana-Farber Cancer Institute/Brigham and Women's Hospital
75 Francis Street, Boston, MA 02115
Email: dbitterman@bwh.harvard.edu
Phone: (857) 215-1489
Fax: (617) 975-0985

## Prior presentations:

None.

## Author Contributions

J.Ga., K.C.K., and D.S.B. conceptualized the study. J.Ga. and K.C.K. performed the interviews. J.Ga. wrote the first draft of the manuscript, and J.Ga., K.C.K., and D.S.B. critically reviewed and edited subsequent versions.M.B., A.C., S.C., P.F.D., S.D., J.Gu., M.J.H., E.H.K., D.E.K., H.J.W.L.A., A.B.L., R.H.M., R.G.M., T.L.N., G.K.S., E.S., B.C.S., U.T., and J.L.W. contributed to study design, interpretation of findings, and critical review and revision of the manuscript.

## Competing Interests

DSB: Associate Editor, JCO Clinical Cancer Informatics (not related to the submitted work), Associate editor of Radiation Oncology of HemOnc.Org (not related to the submitted work) and is on the Scientific Advisory Board of Mercurial AI (not related to the submitted work). DEK: Consultant for AstraZeneca and Genentech/Roche (not related to the submitted work). ABL: Consultant for the Abbott Medical Device Cybersecurity Council (not related to the submitted work). JLW: Editor-in-chief of JCO Clinical Cancer Informatics, consultant for Westat, The Lewin Group, Nemesis Health, UT Medical Branch, ownership of HemOnc.org LLC (not related to the submitted work)

RHM: Advisory Board (ViewRay, AstraZeneca,), Consulting (Varian Medical Systems, Pfizer), Honorarium (Novartis, Springer Nature, American Society of Radiation Oncology), Research Funding (National Institute of Health, ViewRay, AstraZeneca, Siemens Medical Solutions USA, Inc, Varian Medical Systems) - All not related to submitted work.

MJH: Consulting (Replimmune, Deloitte, Guidepoint, GLG)

# Abstract

Large language models (LLMs) integrated into agent-driven workflows hold immense promise for healthcare, yet a significant gap exists between their potential and practical implementation within clinical settings. To address this, we present a practitioner-oriented field manual for deploying generative agents that use electronic health record (EHR) data. This guide is informed by our experience deploying the "irAE-Agent", an automated system to detect immune-related adverse events from clinical notes at Mass General Brigham, and by structured interviews with 21 clinicians, engineers, and informatics leaders involved in the project. Our analysis reveals a critical misalignment in clinical AI development: less than 20% of our effort was dedicated to prompt engineering and model development, while over 80% was consumed by the sociotechnical work of implementation. We distill this effort into five "heavy lifts": data integration, model validation, ensuring economic value, managing system drift, and governance. By providing actionable solutions for each of these challenges, this field manual shifts the focus from algorithmic development to the essential infrastructure and implementation work required to bridge the "valley of death" and successfully translate generative AI from pilot projects into routine clinical care.

# Introduction

Large language models (LLMs) have demonstrated a remarkable ability to interpret complex medical texts, particularly when integrated into agent-driven workflows.[1–3] Their early implementation is guided by statements on the governance of medical machine learning systems and the first wave of position papers on generative AI, which provide valuable guardrails around data privacy, bias, and model evaluation.[4–6] Yet, the existing literature remains largely silent on the day-to-day engineering and organizational work required to implement an LLM into research and clinical workflows. [7–10] Bridging this gap is urgent as new agent-based systems are rapidly being developed. These systems enhance LLMs by enabling them to automate complex procedures; they can execute multi-step plans, use external tools, and interact with digital environments to function as powerful components within larger workflows.[11] While this is promising and hospitals are interested in piloting these systems, their developers still must improvise on how to integrate unstructured clinical data streams, vet outputs, and demonstrate value for overstretched healthcare systems.[12] Clearer guidance could improve the success rate of pilots and help close the last-mile gap in clinical AI.[13]

Here, we address this need by presenting the first practitioner-oriented field manual for deploying generative agents in electronic health record (EHR)-based clinical or research workflows. The field manual distills lessons from organization‑level efforts at Mass General Brigham and an embedded deployment of a task‑bounded immune related adverse event (irAE)‑Agent used for irAE surveillance in patients receiving immune‑checkpoint inhibitors. Immune checkpoint inhibitors have reshaped oncology but introduced immune‑related adverse events (irAEs) that are frequent, heterogeneous, and time‑sensitive.[14–16] Clinical evidence relevant to early recognition and correct CTCAE grading are scattered across problem lists, labs, imaging, and unstructured notes.[17] In busy clinics and emergency rooms, surveillance is episodic and escalation pathways vary, creating room for delays and inconsistent responses. At a high level, the system (i) ingests governed structured and unstructured EHR data, (ii) assembles case‑specific evidence with provenance, (iii) uses an LLM‑based agent to produce a proposed CTCAE grade and supporting rationale, and (iv) routes outputs for human review and sign‑off, with all steps written to an audit log. While developing this system, we worked closely with multi-disciplinary stakeholders, and conducted interviews to better understand facilitators, barriers, and outstanding needs for agent deployment within clinical systems.

The irAE‑Agent provides the running example. Rather than centering model internals, we focus on the sociotechnical path to dependable use in care. Our efforts and interviews revealed five deployment tasks that consumed the majority of effort and ultimately determined success—(i) data integration, (ii) model validation, (iii) economic value, (iv) prompt and model drift, and (v) governance(see Figure 1).

In this article, we describe each of the Heavy Lifts at an organizational- and project-specific level to guide developers, researchers and informatics teams in crossing the last mile of clinical implementation. For each Heavy Lift, we provide specific steps, common failure modes, and artifacts (e.g., evidence‑link requirements, sampling plans for hallucination audits, value workbooks that include oversight OPEX, drift triggers/rollback procedures, and governance RACIs). We aim to complement existing standards and guidelines focused on the models themselves (e.g., TRIPOD/TRIPOD‑AI and related efforts) by providing a practical guide on multi-stakeholder considerations that must be addressed to deploy agentic systems safely and productively in routine care.[18,19] [21]

# Results

Analysis of the interviews revealed a dominant overarching theme: a gap between where the field currently invests its attention and where real-world success is actually determined. While significant attention and effort is devoted to algorithms and models (and advances therein), major efforts in implementation and infrastructure are currently the most challenging and understudied aspects that create a barrier for clinical impact. Below, we describe the 5 major themes that emerged as heavy lifts that are essential to successful deployment of agentic systems within clinical infrastructure (Table 1). We expand on each heavy lift in turn and then synthesise their implications for practice and research in Table 7.

These lifts operate at two levels—organization‑level guardrails and project‑level steps—and they interlock: trustworthy context enables meaningful validation; validation must translate into credible value; value erodes without drift management; and governance binds the system with accountability.
A central observation from our deployment is that the activities that dominate papers and talks—prompt engineering, fine‑tuning, and orchestration—accounted for a minority of total project time. The lion's share was the unglamorous but essential work of data engineering, stakeholder alignment, regulatory navigation, and workflow integration. We describe this as the "80/20 rule" for clinical AI deployment: for every hour spent perfecting a model, expect roughly four hours to make it work in the real world.[22,23]

## Heavy Lift 1: Data Integration

Agentic systems, including the irAE-Agent, are currently largely developed using retrospective datasets, which often means that they are not designed in alignment with data access, representation, standardization, and governance in real clinical settings.[24–26] Integrating EHR data with sufficient latency and security to support HIPAA-compliant generative AI workflows represented our most significant engineering challenge in deploying the iirAE-Agent. We summarize the concrete steps taken at the organizational and project levels, key lessons learned, and fundamental differences between LLM and traditional ML workflows at our institutions.

## Organization Level

At the organizational level, our institution unified longitudinal EHR data into a centralized research enclave built on Snowflake. Automated data transfer pipelines from EHR systems to the research environment were performed and mapped to a common data model for maximizing utility. Specifically, Mass General Brigham uses existing OMOP/i2b2 semantic standard mappings.[27,28] The system employs a standardized daily preprocessing pipeline, which includes splitting unstructured clinical notes into two-line chunks to be stored in data tables using custom tools. This chunking was conducted for storage efficiency and as part of an organizational effort to generate a single note-level embedding service that benefits from consistent sequence lengths. All data transfers occur over Azure private networks with Okta-based SSO controlling access to the research enclave.

Several key lessons emerged at this level. Early investment in a centralized data warehouse provided a foundation for our LLM initiatives, eliminating redundant data extraction and maximizing return on investment (ROI) through a scalable, minimal-preprocessing pipeline. Our cloud infrastructure team enabled these gains by creating predefined "research zones" in Azure. These templates incorporate all necessary security and compliance settings by default, allowing teams to provision new project spaces with speed and confidence. This secure-by-design approach has accelerated project uptake by ensuring a trusted, secure foundation for innovation. When we noticed a persistent gap between clinical needs and technical execution of data pipelines for incoming projects, we established a dedicated clinical data concierge team that provides clinically relevant example queries for the research enclave and supports multiple teams to build reliable clinical business and research cases.

Early on, we found significant interest among clinician innovators in real-time or near-real-time note analysis. However, initial projects using real-time HL7 feeds added complexity and resource requirements before demonstrating proof of value, which stalled effort. We now recommend establishing proof of concept through retrospective or daily batch processing. In fact, while real-time deployment may be appealing, there are a large number of clinical scenarios that can demonstrate clinical value at daily or longer frequencies. Because immediate analysis requires much more project-specific engineering in terms of data integration, it is important to consider whether a use case really requires immediate agent output. This approach makes efficient use of resources and best justifies the higher costs and complexity of real-time implementation using API-first standards like FHIR (Fast Healthcare Interoperability Resources) when it is truly required for a use case.

Furthermore, the choice of platform for data staging and model inference—whether a single-vendor solution (e.g., Databricks, Epic Nebula) or a "best-of-breed" architecture—emerged as an important strategic decision for the success and efficiency of a given project. We recommend carefully considering data staging and inference platforms to balance ease of upfront integration against risk of vendor lock-in. Important considerations here include project urgency and anticipated timescale, clinical scale, anticipated future system upgrades, and latency needs for ongoing deployment. We display the minimum viable infrastructure to replicate similar architectures across three levels in Table 2 & 3.

Notably, LLM workflows differ substantially from traditional ML in this domain. LLM workflows require processing millions of free-text clinical notes, as opposed to smaller structured data queries typical of classical ML. This necessitated entirely new preprocessing steps, such as:  text chunking, vector index creation, and semantic abstraction layers. In contrast, traditional ML projects generally require simple, structured queries of clinical cohort databases.. The rapidly evolving LLM vendor landscape necessitated continuous adaptation, unlike mature ML tool stacks.

At the project level, while we initially planned for real-time inferencing, we shifted our goal to daily monitoring based on deeper consideration of our use case, wherein patients experiencing an irAE need to be registered to the biobank within 96 hours. After discussing the timing needs with the organizational informatics team, we chose to collect data from the established research enclave, which is updated approximately every 24-36 hours. We began by extracting relevant clinical notes and metadata from the research enclave and loading them into secure, project-specific Snowflake sandboxes. These sandboxes are isolated environments preconfigured with the necessary data and compute resources, enabling rapid exploratory data analysis and model testing without the overhead and risk associated with the full production system. The organizational data concierge team assisted us in ensuring that our Snowflake queries were identifying all required data types. Cohort iteration and refinement, as well as preprocessing, were performed within sandboxes. After initial retrospective validation to demonstrate that the agentic system performs above our threshold metrics for deployment (see Heavy Lift 2: Model Validation and Refinement), service account permissions were implemented to ensure that the auditable queries also aligned with Institutional Review Board (IRB) and information security protocols when we transitioned from retrospective to prospective testing. Data transfers from our Snowflake sandbox to API models use Azure private endpoints already established at the organizational level, preventing public network exposure. Template resource groups ensured consistent security settings across cloud storage, compute, web apps, and APIs. Once our data integration pipeline was established, we implemented automatic daily inferencing pipelines through service accounts.

Key lessons at the project level included the following. Working with the organizational informatics team to clarify timing needs for agentic inference allowed us to leverage the research enclave instead of developing custom real-time pipelines, saving significant project resources. Secure sandbox environments enabled rapid cohort and dataset refinement, which revealed important details about note content, clinical note latency, and the impact of preprocessing on data formatting. For example, central preprocessing strips formatting for storage efficiency, affecting downstream rendering for end-users. Vendor tool evolution necessitated infrastructure refactoring approximately every few months, such as transitioning from sequentially chained OpenAI APIs to multi-agent systems capable of parallel execution and heavy concurrent usage.

From an ML comparison perspective at this level, unlike traditional ML, human-in-the-loop annotation workflows became central, requiring budget allocation for API costs. LLM output evaluation necessitated storing model outputs and logging agentic behaviors, more analogous to ML feature importance tracking than traditional model deployment.

## Heavy Lift 2: Model Validation and Refinement

In the development stage of agentic workflows, researchers and developers commonly optimize and validate outside of a real clinical workflow which may not adequately model performance and impact when integrated in a real clinical system. This gap in approaches for real-world evaluation and risk assessment can create challenges for implementation. Iterative validation and refinement served as the linchpin of our irAE-Agent deployment strategy. While this paper focuses on implementation considerations rather than pre-clinical development, we highlight the novel processes required to confirm efficacy and safety in production. Beyond standard retrospective baseline testing and silent prospective runs, our approach required human annotation at each stage, systematic evaluation of model grading outputs and failure modes, hallucination assessment, evidence verification, and controlled human-computer interaction studies.

At the organizational level, following stable data integration, we provisioned secure Snowflake sandboxes where clinical champions could test hypotheses on real data without jeopardizing production systems. Centrally funded "bridge engineers" translated early-priority clinical prototypes into hardened code, reducing integration errors. This was particularly evident with regard to handling missing data, data leakage, and outcome creation. After initial prototyping, each operational project has a "friends-and-family" preview, which extends beyond simple error checking, as LLM/agent applications demand additional red teaming wherein systems are stress-tested against adversarial inputs to ensure smooth, safe adoption.[29,30]

Several key lessons emerged at this level, notably that early sandbox access built institutional trust while surfacing hidden risks at minimal cost. Bridge engineers—developers who can incorporate software engineering best practices, such as unit testing and red teaming—proved critical, eliminating countless handoffs and misspecifications. One de-risking strategy we found to be popular for our LLM projects was a hybrid approach, first using LLMs to extract evidence, then passing these excerpts to conventional classifiers for improved interpretability.

Furthermore, classical ML validation is largely front-loaded and assumes static model behavior. In contrast, monthly LLM API updates transformed validation into a perpetual service. Prompt engineering and qualitative unit tests replaced fixed feature sets. In addition to assessing bias by comparing outputs and performance across demographic categories, the nuanced language in clinical texts used by agents also required us to evaluate how other descriptors in input texts could contribute to bias.[31] For example, we reviewed how social determinants of health cues in notes could impact agent outputs. [32] Failure modes like hallucination and context loss, rare in ML models based on structured data, have become new considerations with LLMs. [33,34]

At the project level, our validation is progressing through four phases. We first established a retrospective gold standard, which took approximately six months of chart curation to create dual-annotated, oncologist-adjudicated notes that provided clear labels, against which prompt engineering could be measured. Gate metrics were prespecified (see Table 4). Zero-shot methods were initially evaluated and failure modes identified in order to break down complicated tasks or common failure modes into a multi-agent system. Ultimately, the final architecture yielded a detection macro F1 of 0.88, sensitivity of 0.84, and precision of 0.95, surpassing our gate thresholds.

After meeting these thresholds, we launched a limited clinical rollout as part of a field study for internal registry curation where clinical research coordinators (CRCs) used the system to assist the curation of retrospective research datasets, accompanied by structured human–computer interaction (HCI) studies. Following positive feedback from this team, the model entered a three-month silent mode; alerts were logged and validated internally by the team but not shared with wider clinicians. The system is now being implemented for our intended use of irAE biobank registration. Given the performance threshold significantly exceeding gate metrics, the system is now also being considered for additional clinical use-cases, including immunotoxicity service triage.

Key lessons at the project level included the following. Strict annotation guidelines and label-writer training were non-negotiable—without high-fidelity labels, we couldn't distinguish model error from label noise. This training included 'cheat sheets' and a library of curated, ambiguous cases to standardize the onboarding process for new annotators. Dual annotation with physician adjudication resolved ambiguities early, helped separate gray cases from absolute errors, and prevented costly relabeling or overoptimizing on noisy labels. This was critical for standardizing complex labeling dimensions such as attribution (e.g., clinician finding vs. patient report) and certainty (e.g., confirmed vs. differential diagnosis). Similarly, agent-level unit tests to evaluate prompt subtask performance—co-written by data scientists and clinical informaticists—drove rapid refinement, while lower-temperature prompts were adopted to improve reproducibility at the cost of some output richness. An incremental rollout mitigated safety risks while providing real-world feedback that was unavailable from static offline tests.

Notably, continuous validation replaced one-off test set paradigms.[35,36] Evaluation metrics expanded beyond classification scores to agreement metrics with prospective CRC labels and survey feedback from field study participants. Robust API management (handling context windows, timeouts, and drift) became integral to validation. Most critically, LLMs exhibited both impressive entity detection and unpredictable hallucinations, for example, attribution of events to immunotherapy when none was mentioned in the note. This required an assessment of not just overall accuracy, but also a more detailed, often manual review of clinical notes to understand and improve agent reliability and fail-safe behavior. Establishing software infrastructure to test multiple LLMs efficiently and effectively allowed us to build methods that reflect state-of-the-art while maintaining safety when new model releases occur.

## Heavy Lift 3: Ensuring Economic Value

he long-term adoption of any clinical AI tool depends as much on credible economics as on technical performance, which should be evaluated throughout the lifecycle of agentic project development.[37] Importantly, while the preclinical performance and agentic inference cost estimations considered by developers are integral to these economics, they are insufficient on their own to justify integration into a workflow. Economic considerations require multi-stakeholder discussions across informatics, developer/researcher, and clinical groups. From the outset, we approached the irAE-Agent's economic value as an, evidence-driven program that advances in parallel with scientific validation.

The organizational approach was to map all candidate use cases to priorities such as: revenue preservation, labor productivity, and quality metric attainment in high-volume services . After identifying labor-intensive tasks, we established a clinical AI sounding board. We centralized large, fixed-cap expenditures (Azure landing-zone hardening, Snowflake pipelines) to maximize access distribution. Variable costs (API calls, Azure Foundry, and elastic GPU usage) were allocated to individual research or collaborative groups. Key templates for research zones, compute, and storage setups ensured default security compliance while streamlining setup times by offloading security engineering (Table 2).

A key lesson was disentangling fixed from variable costs proved indispensable—ROI projections ignoring this distinction collapsed as usage scaled. While LLMs helped with specific use cases, labor substitution was rarely linear: reclaiming 10% of a clinician's time on a task doesn't necessarily equate to 10% cost savings or an increase in patient volume. Business cases needed to emphasize redeployment and task-specific productivity rather than direct replacements. Economic arguments had to target veto holders—typically service line chiefs or finance directors—whose incentives may differ from frontline innovators.

Further, denerative outputs complicate value quantification. Unlike binary classifiers, text summaries introduce subjective quality components (such as clinical relevance, timeliness, and empathy) that resist simple cost-benefit analyses. Each prompt or model upgrade triggers fresh validation cycles, enlarging monitoring budgets. The LLM ecosystem's rapid evolution requires agile procurement strategies to avoid lock-in while maintaining operational stability—a concern that is rarely surfaced in slower-moving, structured data-based ML projects.

At the project level, ROI modelling proceeded along two axes. The "hard" axis captured marginal savings from reduced manual chart abstraction for research—historically a six‑figure annual research expense—while the "soft" axis tracked safety gains, clinical research coordinator experience, and research acceleration. A live dashboard reported total notes evaluated, inference cost, and processing latency for each day, reinforcing the narrative that unit economics improve as volume scales.

By pairing our own research funds—earmarked for model development and validation—with the hospital's existing cloud and data warehouse stack, we stretched a modest budget into a production-ready prototype. This "use‑what‑we‑have" strategy enabled streamlined costs and related institutional approval that pure research dollars alone could not have achieved.

We first validated a lightweight, on-premises model to prove feasibility, then graduated to larger open-source and API models as performance targets increased. Cost tracking across each upgrade and prompt engineering cycle now puts inference at roughly $2 per 100 clinical notes using the irAE-agent.

As we scale, we'll benchmark the time saved by reviewers and residual error rates to inform the next round of optimization and cost-effective LLM selection.

From an ML comparison perspective at this level because unstructured text output is not automatically mappable to categorical events, value realization needs to consider workflow timeliness and safety in addition to accuracy. Monitoring overhead—auditing hallucinations, drift, and context loss— and assessing system performance as LLMs advance became a necessary operational expenditure line item at a scale not encountered with traditional ML models.

## Heavy Lift 4: Managing Model Drift and Data Drift

The rapid rise of agentic AI built on generative LLMs has outpaced the development of robust post-deployment monitoring methods.[35,38–42] Unlike traditional machine learning, there are currently no widely accepted or validated approaches for detecting model or data drift in these systems. As a result, researchers and developers must direct innovation not only toward pre-clinical performance optimization, but also toward creating validated, actionable monitoring strategies and infrastructure that can operate alongside the agentic system.

At the organizational level, Hospital leadership aimed for always-on monitoring: every production LLM must collect automated statistics on usage and performance compared to reference sets or gold standards. While no full protocol currently exists for specific metrics, ongoing discussions with the AI team aim to establish best practices and develop new monitoring methods and tools.

The need for deep expertise led to the formation of a dedicated AI governance/clinical AI team. The team emphasized "continuous, not quarterly" surveillance through research methods and feedback loops. Human validation surfaces clinical subtleties that automated dashboards miss: while statistics provide valuable insights, they cannot replace human validation and manual data curation.

There are many similarities in the need for ongoing monitoring of ML and LLM models in the clinical setting. However, both approaches require ground truths, which remain challenging in clinical settings, and involve non-trivial clinician time and financial investment. New problems that emerge for organisations in the LLM setting are centered around the organisation's monitoring of API usage. For example, there is a new potential for jailbreaks or changes in external provider API models, which can impact multiple projects and departments.

At the project level, technical surveillance operates on two orthogonal axes, addressing both drift (behavioral changes that occur without altering the model version or input schema) and shift (behavioral changes resulting from external alterations to inputs or model artifacts). For model drift, we propose weekly re-scoring against frozen gold-labeled test sets with longitudinal performance tracking (Table 5). API endpoints will be pinned to exact model versions to isolate vendor upgrades from true performance erosion. For data drift, incoming clinical notes will be embedded and compared with historical distributions, with additional stress-testing through synthetic reformatting to simulate new EHR templates and transcription formats. Beyond these axes, we are evaluating the efficacy of direct LLM-as-judge monitoring of the LLM's live output behavior, including automated tracking of failure modes such as factual consistency, tone, and hallucination rates on production inferences.

We learned that while monitoring input data and test set performance is necessary, it is insufficient for generative agents; direct surveillance of output behavior is required to catch emergent failure modes not present in static datasets. These methods are currently being implemented, with evaluation planned as sufficient data accumulates. Scalable drift evaluation remains challenging, as it requires a balance between manual review and automated testing approaches. The proposed framework emphasizes early detection through multiple complementary metrics, rather than relying on any single indicator.

Regarding differences, drift manifests as context loss or hallucination rather than simple metric degradation. Remediaton often involves prompt engineering rather than complete LLM retraining. Akin to hyperparameter tuning of ML models, which is performed far less frequently, the need for continuous prompt and version management becomes a standing operational cost, foreign to many ML projects. However, deciding when to rebuild the agentic architecture or if fine-tuning is needed with additional data, rather than iterative prompt engineering, is use-case specific.

## Heavy Lift 5: Governance

Deploying a generative AI agent in direct patient care raises questions of ethics, accountability, and regulatory compliance that extend well beyond traditional supervised learning projects.[43] Developing and agentic system without upfront consideration of the downstream governance requirements could lead to a system that has demonstrated good performance, but insufficient evidence and embedded infrastructure for safe, ethical, and acceptable clinical translation. To translate the irAE-Agent from prototype to clinical tool, we distinguished organization-level guardrails—common to any hospital LLM deployment—from project-specific controls tailored to irAE surveillance.

Hospital leadership convened an enterprise AI governance board including representatives from clinical, legal, security, patient experience, and finance. The board mandated lifecycle checkpoints—purpose definition, safety, efficacy, effectiveness, and ongoing surveillance—for every generative AI service. A central policy requires each checkpoint to be documented in a project-charter responsibility, accountability, consultation, and information (RACI) matrix [44], ensuring that responsibility, accountability, consultation, and information flows are explicitly defined before work begins.

Institutional "AI readiness" governs project velocity: hospitals with codified multidisciplinary oversight and version-pinning practices can apply existing guardrails to new use-cases, whereas green-field institutions must build policy in parallel with technical work. Monthly or ad-hoc meeting cadence proved superior to traditional quarterly board cycles, matching the tempo of foundation-model releases without sacrificing due diligence. Clarifying for the IRB that prompting does not expose PHI for open-source LLMs behind the institutional firewall or HIPAA-compliant API LLMs helped the IRB to make a waiver of consent determination. While monitoring methods are often developed at the project level, organizational-level monitoring proves critical for ensuring accountability and safety.

The open-ended, unstructured nature of LLM inputs introduces jailbreak and prompt-injection risks unseen in traditional ML, making red-teaming and continuous content-safety audits mandatory. Silent vendor upgrades can invalidate prior safety evidence, requiring agile re-validation, whereas traditional models remain static once deployed. Prompt engineering, where real clinical data are used to optimize prompting strategies, adds a new privacy consideration for governance and regulatory bodies, and can create confusion with regard to the presence or absence of PHI privacy risk. With traditional ML, optimizing performance requires training/tuning on real clinical data, and (depending on the architecture) PHI may reside in the learned weights. Conversely, agentic system performance may be optimized via prompt engineering on real clinical data, but if the prompts are free of PHI, the resulting optimized system does not necessarily contain PHI. Governance approaches need to carefully delineate and consider privacy risk as it relates to sharing clinical data, LLMs trained/tuned on clinical data, and prompts optimized on clinical data.

We established a project-level governance structure to complement the organizational governance structure, spanning oncology, informatics, and engineering. We mapped governance goals to five lifecycle phases and codified accountability through a RACI decision matrix (Table 6).

The RACI framework clarified roles across key decisions: defining purpose and operational lifetime was the responsibility and accountability of the oncology principal investigator; data type selection required broader clinical team responsibility with data science and engineering consultation; infrastructure decisions like hosting platform required engineering responsibility with Cloud Operations consulting and principal investigator consultation. By pre-allocating these roles in the project charter, we eliminated ad-hoc email chains that could stall AI deployments.

Several key lessons emerged at this level. Early governance integration averted late-stage friction and accelerated approvals. Explicit instruction to reviewers that prompts never contain PHI, whereas fine-tuned models might, shortened IRB deliberations. The crawl-walk-run rollout surfaced usability barriers, safety issues, and end-user concerns long before the full release. Real-time dashboards linked cost and latency to safety metrics, preventing performance “fixes” that would have resulted in budget overruns.

Unlike fixed ML models, LLM agents require governance that evolves with each prompt revision or vendor weight refresh. Real-time safety and integrity dashboards became the primary assurance mechanism. The attack surface widens significantly compared to ML: jailbreaks and prompt injections can repurpose narrowly scoped agents into unrestricted text generators, requiring continuous red-teaming rather than periodic model-drift audits.

Table 7 summarises how each heavy lift exposes a mismatch between the way model and algorithm builders typically conceptualise the problem and the sociotechnical work that ultimately determines successful deployment. It also distils the resulting implications for project leaders, organisational leaders, and academic researchers and model developers, making explicit the practice and knowledge agenda that emerges from the irAE-Agent deployment

# Discussion

This work makes several concrete contributions to the emerging practice of deploying generative AI agents in clinical care. First, we organize implementation into five Heavy Lifts—data integration, model validation and refinement, ensuring economic value, managing model and data drift, and governance—and for each we provide organization- and project-level key steps, key lessons, and key differences from traditional machine learning. Together, these structured resources function as an institutional field manual that health systems can adapt rather than starting from first principles. Second, we pair this framework with end-to-end infrastructural blueprints, including example infrastructure tiers for clinical decision support and LLM applications, as well as an internally used feasibility mapping between project types and infrastructure tiers, gate metrics for advancing between deployment stages, and monitoring approaches for hallucinations, drift, and context loss. Third, we articulate governance artifacts such as a RACI matrix for deployed systems, clarifying roles and accountability for clinical, technical, and operational stakeholders. Collectively, these tools provide templates that can be repurposed and localized by other institutions seeking to move from pilots to durable, accountable deployments.

Our deployment of the irAE-Agent demonstrates that integrating generative AI into clinical practice is fundamentally a sociotechnical challenge, not merely a technological one. Navigating hurdles in data integration, model validation, economics, governance, and monitoring required aligning diverse stakeholders, from clinicians to administrators, and building trust through deliberate, continuous validation. This effort laid the groundwork for the successful deployment of an agent-as-copilot, effectively scanning patient data to provide early alerts for irAEs, thus enabling proactive detection and intervention. Our experience provides a practical roadmap for other institutions, demonstrating that the clinical impact of generative AI hinges not on algorithmic sophistication alone, but on careful preparation, proactive governance, and iterative refinement.

# Methods

Following the development and initial deployment of the irAE-Agent to address a clinical and research need (described above), we observed that "last-mile" implementation raised complex technical, operational, and clinical challenges and that existing literature offered limited practical guidance. We therefore conducted a qualitative study to characterise the development and implementation issues that mattered most in practice. Between June 2025 and August 2025, we conducted a semi-structured interview study with 21 technical, operational, and clinical stakeholders who were directly involved in the development and implementation of the IrAE-Agent at a project level or an organizational level. The primary objective of the interviews was to gain a comprehensive understanding of the development and implementation issues that mattered most in practice. This study was approved by the Institutional Review Board of the Massachusetts Institute of Technology, Project E6632.

## irAE-Agent System Overview

To ground the qualitative findings in a concrete implementation, we deployed an automated irAE surveillance agent that screens clinical notes for patients treated with immune checkpoint inhibitors and produces evidence-linked, CTCAE-anchored outputs for human verification (Figure 2).

While the goal of this field guide is to provide practical guidance on implementation, not on the technical development and validations of irAE-Agent, here we provide details of the irAE-Agent to ground the project-level discussions. Full description of the irAE-Agent methods and performance is reported separately, and an overview of the system itself can be found in Figure 2. The irAE-Agent is an automated surveillance system that screens clinical notes daily to detect irAEs in patients treated with immune checkpoint inhibitors within the prior year. irAEs are unpredictable, multi-organ toxicities that can emerge weeks to months after therapy, and their signs and symptoms are usually documented in free-text notes, making timely, consistent detection exceptionally challenging.(22–24)

Each morning, the organizational data pipeline at Mass General Brigham pulls all information from the Epic EHR system and moves this to an institutional Snowflake research enclave (organizational pipelines are described in more detail in Heavy Lift 1: Data Integration). Subsequently, our project-level pipeline extracts a targeted cohort of progress notes, discharge summaries, and oncology documentation for patients who have received at least 1 approved immune checkpoint inhibitor infusion within the prior 12 months based on infusion records at our institution. The project workflow is directed by serverless applications, which are cloud-based programs that automatically run tasks on demand without requiring manual server management. These applications handle daily cohort creation and LLM-based extraction of irAEs, along with model explanations and evidence retrieval. The applications also automate report generation and display the extracted notes and model findings for human verification, which can be reviewed offline and behind the organizational firewall. The system currently interfaces with the OpenAI API through HIPAA-compliant endpoints.

The pipeline is completed within 30 minutes each morning automatically, generating structured reports that are routed to the research team. Model, data, cost, and token logging at each stage enables continuous monitoring for drift and safety issues to be reviewed as new methods are evaluated for automating the monitoring of LLMs, as detailed in subsequent sections. This architecture and quality assurance protocol were purpose-built to strike a balance between performance requirements and stringent healthcare data governance standards.

Developing and implementing this irAE-Agent within the Mass General Brigham information ecosystem required input and collaboration across research and operational informatics/data science, regulatory, clinical, and governance teams. This process surfaced challenges likely to be shared across others interested in deploying generative agents that use EHR data, motivating the semi-structured interviews described below in order to gain deeper insights into existing challenges and strategies for agentic implementations for clinical workflows. Findings from these interviews and our experience which we organize into five “heavy lifts”. With this overview provided for additional context, we now pivot to a description of the five “heavy lifts”.

## Participant Recruitment

We used purposive sampling to gather insights from key stakeholders involved in developing and implementing the irAE-Agent. Twenty-one stakeholders were recruited from across the organization. One stakeholder was on maternity leave, so declined. All other stakeholders accepted the interview. On average, participants had 18 years of experience and brought technical (38%), operational (38%), and clinical (24%) expertise (Supplementary Table 1). Technical stakeholders included technology developers and IT infrastructure managers; operational stakeholders included clinical research coordinators who were the end users of the model outputs, and senior managers and directors of the implementation of emerging technologies in clinical practice; and clinical stakeholders included frontline clinicians. To mitigate potential bias, we ensured diverse representation across technical, operational, and clinical domains, organizational levels, and years of experience.

## Interview Guideline Development and Validation

One author (J.G.) reviewed meeting notes from meetings about the development and implementation of the irAE agent held between 1st July date and 15th August 2025. Based on the insights obtained, a preliminary semi-structured interview guide was developed. Another author (K.K.) conducted pilot interviews with four technical, operational, and clinical experts (different from the 21 interview participants in the study) to test and refine the questions for the semi-structured interviews. The final semi-structured interview guide focused on: key steps in development and implementation of the irAE agent, key lessons learned, and key differences versus traditional machine learning—at both the project and organizational levels.

### Data Collection

Two authors (K.K. and J.G.) conducted one-on-one, semi-structured, in-depth interviews with 21 participants. K.K. is a trained qualitative researcher with experience in the development and implementation of AI solutions in clinical care. All interviews were conducted via Zoom and lasted for 30 minutes to an hour. The interviews were recorded and later transcribed for analysis via an automated transcription software. All participants provided verbal informed consent to participate in interviews, to have the interviews recorded, and to allow anonymized data to be used in qualitative analysis.

### Data Analysis

To analyze the transcripts and extract key insights, we used an inductive coding process. We manually coded the transcripts using a constant comparative method. This process began with initial code identification through iterative analysis, followed by grouping these codes into overarching themes that captured key dimensions of the IAre agent development and implementation. A third author (D.B.) independently reviewed the coding framework. Minor discrepancies were resolved through consensus. Five key themes emerged from the data analysis—(i) data integration, (ii) model validation, (iii) economic value, (iv) prompt and model drift, and (v) governance—that consumed the majority of effort and ultimately determined success (see Figure 1). All authors reviewed and approved the final analysis.

### Data Availability

The qualitative data generated and analyzed during this study are not publicly available due to ethical and confidentiality considerations. Participants did not consent to public data sharing. Details of the experience and demographics of all interview participants are available in Supplementary Table 1.

## Acknowledgements

The authors acknowledge financial support from the National Institutes of Health National Cancer Institute (U54CA274516-01A1 [J.G., D.S.B, D.E.K., G.K.S.], R01CA294033-01 [J.G., P.F.D., J.L.W, M.J.H., E.S., D.E.K., G.K.S., D.S.B]), the American Cancer Society and American Society for Radiation Oncology, ASTRO-CSDG-24-1244514-01-CTPS Grant DOI #: https://doi.org/10.53354/ACS.ASTRO-CSDG-24-1244514-01-CTPS.pc.gr.222210 [D.S.B.], a Patient-Centered Outcomes Research Institute (PCORI) Project Program Award (ME-2024C2-37484) [D.S.B.], and the Woods Foundation [D.S.B.]. All statements in this report, including its findings and conclusions, are solely those of the authors and do not necessarily represent the views of the Patient-Centered Outcomes Research Institute (PCORI), its Board of Governors or Methodology Committee.

## Figure Legends

**Figure 1:** Two-tier framework illustrating the relationship between organizational infrastructure investments (bottom tier) and project-specific implementations (top tier) for deploying a generative AI agent for immune-related adverse event (IRAE) detection at Mass General Brigham. The five project-level "heavy lifts" (numbered 1–5) build upon corresponding organizational foundations, with arrows indicating dependencies. Each heavy lift corresponds to a deployment phase and includes specific activities and measurable outcomes. This framework demonstrates how institutional readiness enables rapid clinical AI deployment.

**Figure 2:** System architecture for automated irAE detection and reporting. The daily workflow begins with on-premise data processing, where (1) all MGB clinical notes are extracted from the EHR, processed, and (2) stored in the MGB Snowflake Research Enclave. Within a secure MGB Azure tenant, (3) a 5 AM trigger initiates an Azure Function Orchestrator that (4) deploys an extractor container app to fetch new ICI patient notes. (5) These processed notes are stored in Azure Blob Storage via a private endpoint. (6) A second predictor container app is then triggered, (7) leveraging a private Azure OpenAI endpoint to predict irAEs from the notes. (8) The final predictions and a summary report are written to Blob Storage. These are then reviewed by our research team for eligibility in biobank recruitment and other use-cases. MGB, Mass General Brigham; EHR, electronic health record; ICI, immune checkpoint inhibitor; irAE, immune-related adverse event.

## Table Legends

**Table 1:** The five “heavy‑lift” domains for irAE agent implementation (column 1), with organization‑ and project‑level steps (columns 2–3, 5–6), lessons learned (columns 4, 7), and key differences from classic machine learning workflows (columns 4, 7).

**Table 2:** Key infrastructure tiers for deploying clinical decision support large language models (CDS‑LLMs): column 1 lists core infrastructure components; columns 2–4 outline the capabilities, launch timelines, infrastructure spend, and team composition across foundational, sustainable, and scaled deployment tiers.

**Table 3)** Feasibility of representative CDS‑LLM use cases across infrastructure tiers: rows reflect common patient-, care team-, research-, and employee-facing scenarios; columns indicate feasibility at each infrastructure level (High = fully feasible; Med = possible with trade‑offs; Low = not advised). Use cases are illustrative, not exhaustive.

**Table 4:** Prespecified operational metrics (“gate” criteria) for irAE agent deployment. Columns list each metric, the minimum threshold required for deployment, observed performance over a 12-week prospective study (N = 12 weeks), and the rationale for its inclusion.

**Table 5:** Proposed approaches for monitoring agentic clinical systems in dynamic healthcare environments. Rows outline recommended methods, alert thresholds, and evaluation frequencies for detecting and addressing data drift, model drift, input or population shifts, and model release effects. Metrics include embedding and TF-IDF divergence, gold-set rescoring, inter-model disagreement, and simulation-based shift testing. Criteria focus on safety, reproducibility, and operational integrity.

**Table 6** Project-level governance decision matrix for agentic system deployment. Rows detail each deployment phase (security & legal, safety, efficacy, effectiveness, and surveillance) and their associated core questions, key decisions, and responsible (R), accountable (A), consulted (C), and informed (I) stakeholders. Matrix clarifies oversight, roles, and escalation pathways at each phase.

**Table 7:** Implications of the five heavy lifts for clinical LLM agent deployment. For each heavy lift, the table contrasts how model and algorithm builders commonly conceptualise the issue with the implementation realities observed in our deployment, and summarises implications for project leaders, organisational leaders, and academic researchers and model developers.

# Organisational Framework for Clinical LLM Agent Deployment

Mass General Brigham IrAE Agent Implementation

## PROJECT LEVEL

### 1 Data Integration

Phase 0: Security/Privacy

- Cross Hospital Unification
- IrAE Specific ETL
- Cohort definition
- Daily streaming

2-week Proof of Concept

### 2 Model Validation

Phase 1: Assessment

- Gold standard curation
- Silent deployment
- Clinical Validation
- HCI Optimization

F1: >0.75, Sens: >0.75

### 3 Economic Value

Phase 2-3: Deployment

- ROI metrics
- Cost-benefit analysis
- Efficiency tracking
- Value communication

50% review efficiency

### 4 Drift Management

Phase 4: Monitoring

- Proposed metrics & methods
- Weekly evaluation
- Prompt engineering
- Version control
- Embedding analysis

<5% drift tolerance

### 5 Governance

All Phases

- Committee oversight
- Phase-gate review
- Accountability matrix
- Ethical compliance

Full Compliance

## ORGANIZATIONAL LEVEL

### DATA INFRASTRUCTURE

- Snowflake warehouse
- Epic Integration
- Medallion architecture
- Research Sandboxes
- Automated ETL

Multi Project reuse

### SECURITY & COMPLIANCE

- HIPAA OpenAI API
- Azure endpoints
- BAA Agreements
- PHI Governance
- Audit logging

Enforced Safety

### ECONOMIC FRAMEWORK

- Infrastructure budget
- ROI tracking
- Cost models
- Priority alignment
- Value metrics

Hard Milestones

### MONITORING

- MLOps platform
- Drift detection
- Dashboards
- Alert systems
- Evaluation tools

Clear Review

### GOVERNANCE

- VP-level Committee
- AI policies
- Review cycles
- Stakeholder board
- Ethics framework

Streamlined Acceptance

Project-specific implementations (IrAE-Agent) anizational Infrastructure Investments

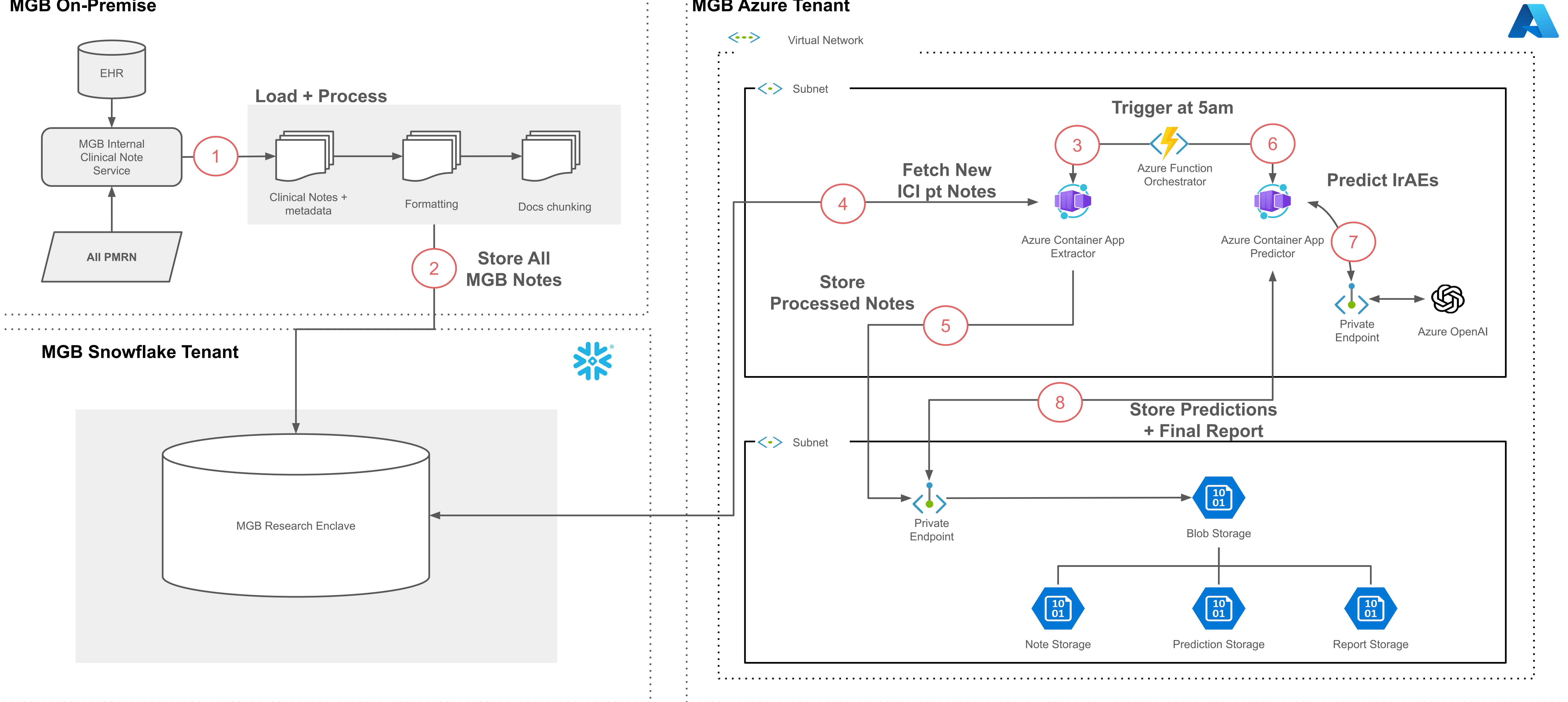

**Table 1: Overview of irAE Agent Heavy Lifts at Organization and Project Level**

| Heavy-lift item | Organization — key steps taken | Organization — key lessons learned | Organization — key differences vs. classic ML | Project — key steps taken | Project — key lessons learned | Project — key differences vs. classic ML |
|---|---|---|---|---|---|---|
| **1 Data Integration** | • Unified longitudinal EHRs from hospitals into Snowflake research zones<br>• Daily Epic → Snowflake ETL (bronze → silver → gold) with in-house note-parsing tools<br>• Azure private networking & service-account RBAC<br>• “Data concierge” translates clinical asks to SQL bridge technical gap<br>• Leveraged existing OMOP/i2b2 models + pre-hardened Azure/OpenAI landing zones | • Up-front warehouse spending accelerates LLM work<br>• Start with batch feeds; real-time ≈ 10 % extra value for very high cost<br>• Security/API tickets may take several months → submit Day 1<br>• Re-use queries and HL7 pipelines / avoid duplication<br>• Precise data-element specs speed approvals & clinician clarity in requests<br>• Existing enterprise cloud templates offload security engineering | • Millions of notes vs. thousands of rows<br>• Different preprocessing of notes, ie, chunking vs structured<br>• Near-real-time note pipelines vs. monthly SQL dumps<br>• Free-text PHI needs masking/granular entitlements + audit logs<br>• Agents need a semantic layer; ML could hit raw tables<br>• Weekly vendor/tool churn vs. stable ML stack<br>• Difficulty de-identifying notes → secure enclaves, not CSV exports | • Extracted irAE-relevant notes + metadata into secure sandboxes<br>• Service-account governance tied to IRB approvals for repeatable queries<br>• Data moved from sandbox → compute via private endpoints | • Concierge support bridged the clinician–engineer gap<br>• Proof-of-value before latency optimisation<br>• Precise cohorts reduced downstream label noise | • Needed annotation interface<br>• Continuous vendor/tool churn; infra refactor every few months, e.g., OpenAI versions vs langchain vs OpenAI agents, SGLang vs VLLM vs Ollama etc |
| **2 Model Validation & Refinement** | • Secure sandbox for hypothesis testing<br>• Funded “bridge” engineers for priority projects | • Early sandbox builds trust & surfaces risks cheaply<br>• Bridge engineer cuts integration errors ×10<br>• Interpretability layer by using extract with LLM → classic ML → build clinician trust<br>• Have “Friends-and-family” preview for new units before go-live | • Validation is continuous; API versions roll out monthly<br>• Prompt engineering + qualitative tests replace static feature sets<br>• Must justify outputs & validate evidence & block bias (e.g., SDoH leakage) | • Collect baseline-then-prospective loop (retrospective gold set → silent mode → controlled rollout)<br>• 4-phase cycle: baseline (F1 0.89); 3-mo silent mode ; limited field study rollout; HCI studies<br>• Prompt unit tests co-written by clinicians | • Strict annotation guidelines + labeler training indispensable + dual annotation<br>• Lower-temperature prompts ↑ reproducibility but may impact structured outputs<br>• Incremental rollout manages risk & gathers feedback<br>• Setting realistic user expectations on model processing times.<br>• Create a model card to transparently demonstrate the model's capabilities to users | • Failure modes include hallucination/context loss— not seen as concretely or as convincingly in traditional ML |
| **3 Ensuring Economic Value** | ● Aligned use cases to institutional priorities (revenue, labour, quality)<br>● Hard- & soft-ROI model; fixed vs. variable ledger separation | ● ROI falls apart without a CapEx/OpEx split<br>● Labor savings ≠ head-count cuts; emphasize redeployment<br>● Pitch economics to veto-holders (service-line chiefs) | ● Generative outputs subjective → harder value quantification<br>● Each prompt/model upgrade incurs fresh validation cost<br>● Fast tech half-life demands agile procurement | ● Leveraged existing Azure K8s & Snowflake (no duplicate audits)<br>● ROI model on dual axes (cost-savings & safety/ research)<br>● Tracked per-note inference cost, GPU cost, and time | ● Re-use of existing resources enables frugality<br>● Unit economics improve with volume; show this early | ● Textual outcomes are not binary; they include timeliness & safety value<br>● Monitoring overhead is a material part of OpEx |
| **4 Model / Data Drift** | ● Continuous monitoring: daily human audit + stats<br>● Weekly rescoring on frozen gold set; API version pinning | ● Human audit catches nuances that dashboards miss<br>● Separate model vs. data-drift for fast RCA<br>● Vendor version-locking non-negotiable | ● Stochastic outputs; tiny prompt or vendor change shifts behaviour<br>● Requires always-on surveillance vs. quarterly ML audits | ● Create a pipeline to collect information from the input and model outputs<br>● Embed/TF-IDF shift checks; template- & transcript-shift stress tests<br>● Staged remediation: re-prompt → replay → governance sign-off<br>● Shadow-run cheaper models before swap-out | ● Cost/latency logged with accuracy to weigh fixes<br>● Criteria to trigger reprocessing/retraining due to model improvements or updates | ● Drift shows as hallucination or context loss, not just metric drop<br>● Must track hallucination / unsafe-content flags |
| **5 Governance** | • Multidisciplinary committee mirrors enterprise board<br>• Lifecycle checkpoints: Purpose → Safety → Efficacy → Effectiveness → Surveillance<br>• RACI matrix in the project charter | • AI readiness sets velocity ceiling<br>• Monthly/ad-hoc cadence beats quarterly boards | • Conversational outputs need red-teaming for jailbreaks<br>• Silent vendor upgrades can void safety evidence; agile re-validation is required | • Phase-0 IRB & CDS classification<br>• Hazard analysis with adversarial document perturbations<br>• Crawl-walk-run rollout to trusted users | • Governance embedded Day 0 prevents later friction | • Governance must evolve with each model refresh and layers with agents; ML often “set-and-forget”<br>• Needs real-time safety & integrity dashboards, not just periodic drift audits. |

AI: Artificial Intelligence; API: Application Programming Interface; CapEx: Capital Expenditure; CDS: Clinical Decision Support; CSV: Comma-Separated Values; EHR: Electronic Health Record; ETL: Extract, Transform, Load; F1: F1 Score (a measure of a model's performance); GPU: Graphics Processing Unit; HCI: Human-Computer Interaction; HL7: Health Level Seven (set of international standards for electronic health information exchange); i2b2: Informatics for Integrating Biology and the Bedside; IRB: Institutional Review Board; IrAE: Immune-related Adverse Event; K8s: Kubernetes; LLM: Large Language Model; ML: Machine Learning; OMOP: Observational Medical Outcomes Partnership common data model(a global standard for transforming disparate healthcare data); OpEx: Operating Expenditure; PHI: Protected Health Information; RACI: Responsible, Accountable, Consulted, and Informed; RBAC: Role-Based Access Control; RCA: Root Cause Analysis; ROI: Return on Investment; SDoH: Social Determinants of Health; SQL: Structured Query Language; TF-IDF: Term Frequency-Inverse Document Frequency.

**Table 2. End-to-end Infrastructure Tiers for CDS-LLMs (capabilities, time-to-launch, and cost)**

| Component | Foundational (MVI for day-to-day ops) | Sustainable | Scale |
|---|---|---|---|
| **Overview** | Initial setup. Characterized by limited compute (e.g., a single GPU), manual data pulls, and basic security (BAAs, private network). Suitable for low-risk, retrospective projects. | Growing capability. Involves departmental compute clusters, some real-time data pipelines, and template-based security policies. Can support validated operational tools and pilots for higher-risk applications | Mature, enterprise-level system. Features robust, scalable compute (cloud APIs, in-house models), automated observability, and deeply integrated, role-based governance. Can handle high-risk, real-time, and patient-facing applications. |
| **Data access & storage** | Automated ETL from ≥ 1 source into *single-tenant* HIPAA-compliant cloud warehouse; nightly batch cadence | Near-real-time sync from full EHR (notes + labs + imaging); multi-source unification | Real-time FHIR / event streaming; multi-site federation; on-demand synthetic datasets |
| **Compute & model hosting** | Private VNET; dedicated GPU instance(s) or on-prem GPU; manual scaling | Auto-scaling clusters in one region; mix of commercial APIs and self-hosted models | Multi-region, global load-balancing; infra optimised for large-scale training & SFT |
| **LLM / agent capability** | 1 commercial model (BAA) + prompt templates | Prompt / model A-B tests; gateway to several OSS & commercial models | Tool-calling agentic workflows; in-house model R&D and fine-tunes |
| **Security & governance** | Responsible-AI Committee; RBAC; subnet isolation; baseline policies | Automated compliance & usage dashboards; human-in-the-loop tiers | Zero-trust, advanced threat detection, reusable CI/CD templates |
| **Monitoring/ observability** | Cost log + manual output review | Real-time dashboards, drift alerts | Predictive analytics, auto-remediation, BI roll-ups |
| **Spin-up time (typical)** | ≤ 6 weeks | ≈ 3 months | 6 + months |

| **Infra Spend**<br>*On-Premise / Self-Hosted* | Incl. initial server purchase ($30k–$100k, amortized) + annual power, cooling, & rack rental ($10k–$20k) | Maintenance, power, and colocation for a small cluster ($40k–$100k) | Significant costs for large cluster maintenance, power, and data center space ($100k–$300k+) |
|---|---|---|---|
| **Infra Spend**<br>*Cloud & API Services* | Cloud VM/GPU rental, data storage, and moderate API inference spend ($50k–$150k) | Higher volume API inference, managed databases, and auto-scaling compute ($200k–$500k) | Extensive API usage, multi-region cloud services, and specialized infra for training/SFT ($500k–$1.2M+) |
| **Core FTEs & Functions** | ● **AI Leader / PI** (0.1–0.2 FTE)<br>● Cloud / Infra architect<br>● Project-dedicated ML / AI engineer (0.5-1 FTE)<br>● Security & Compliance engineer (shared)<br>● Clinical informatics liaison (shared)<br>● **Legal / Risk counsel** (0.1 FTE shared or external)<br>**Total ≈ 3–5 FTE** | ● Foundational roles plus:<br>● DevOps / SRE<br>● API & Integration engineer<br>● UI/UX / Front-end dev<br>● Digital literacy<br>● MARCOM / Adoption lead (0.5 FTE)<br><br>**Total ≈ 6–10 FTE** | ● Full in-house AI department:<br>● Leadership & product owners<br>● Multiple ML/AI engineers<br>● DevOps + 24 × 7 SRE<br>● Data & Analytics team<br>● Security Ops & Threat Intel<br>● Bio-ethics & Legal team<br>● Digital literacy<br>● MARCOM & Change-management<br>● API & Integration squad<br>**Total > 10 FTE** |
| **Salary pool (fully-loaded)** | **$450 k – $800 k** | **$900 k – $1.8 M** | **$2 M +** |
| **Total annual run-rate**<br>(infra + people) | **≈ $0.6 M – $1 M** | **≈ $1.2 M – $2.4 M** | **≥ $2.6 M** |

**Table 3. Project-type × Infrastructure-tier feasibility**

| Use Case* | Foundational | Sustainable | Scale |
|---|---|---|---|
| **Patient-Facing** | | | |
| Patient Interactive Chatbot | Low | Med | High |
| Automated Portal Messaging | Low | Med | High |
| Education/Literacy/Translation Services | Med | High | High |
| **Care Team (Clinical & Operational)** | | | |
| Real-time Clinical Decision Support (CDS) | Low | Med | High |
| AI-Assisted Quality Assurance | Med | High | High |
| Inpatient / Patient Summaries | Low | High | High |
| Tumor Board Prep | Low | Med | High |
| Message Triage / InBasket Drafts | Med | High | High |
| Revenue Cycle (Prior Auth, Denials, Coding) | Med | High | High |
| General Agentic Clinical Workflows | Low | Med | High |
| General Agentic Operational Workflows | Med | High | High |
| **Research** | | | |
| Retrospective Chart Review & Summarization | High | High | High |
| Research Admin (Grant Writing/Editing) | High | High | High |
| Multi-site Observational Study | Med | High | High |
| **Employees** | | | |
| Employee-Facing Interactive Chatbot | Low | High | High |
| AI Text Editing Assistant | Low | High | High |
| High = fully feasible; Med = possible with trade-offs; Low = not advised. | | | |

*Use cases are not intended to be inclusive of all possible agentic use-cases, but are instead selected from common examples encountered in the course of this work and described in the literature.

**Table 4: Predefined IrAE-Agent Operationalization Gate Metrics**

| Metric | Threshold | Achieved (N=12 weeks) | Rationale |
|---|---|---|---|
| Macro F1 Detection of any IrAE | ≥ 0.75 | 0.91 | Patient safety when implemented with a human-in-the-loop |
| Precision | ≥ 0.80 | 0.92 | Alert-fatigue control |
| Median time-to-alert | < 2 days | 24 hrs | Workflow relevance |
| Override rate | < 10% | 2% | Usability, intended use with human-in-the-loop |

The override rate was calculated as the percentage of label disagreements observed in the prospective setting when a human annotator reviewed the screened notes. This differs from accuracy because human labels can introduce subjectivity in borderline cases, which may be considered incorrect. However, the intended use of the system is with humans in the loop, making this a suitable end-user metric.

**Table 5: Proposed methods to monitor agentic systems in dynamic health systems**

| | Data | Model |
|---|---|---|
| **Drift**<br>Behaviour changes *without* altering the model version/input schema | **Example:** Changes in subpopulation demographics served in the hospital<br><br>**Methods**<br>● **Embedding & TF-IDF divergence**: Cosine-distance vs. baseline centroid & KL-divergence on top-K tokens<br>● **Note-type mix** (OP, IP, ED)<br>● **Average note length** and template ID distribution<br><br>**Frequency:** daily snapshot, formal weekly report<br><br>**Proposed alert threshold:** 2 s.d. shift in any metric for ≥2 consecutive weeks | **Example:** API model micro updates, agent routing, and structured output changes.<br><br>**Methods**<br>● **Gold-set F1 & AUROC** re-scored weekly<br>● **Inter-model disagreement rate** (production vs. pinned GPT-4 reference)<br>● **Human agreement** on rolling 20-note daily sample<br><br>**Frequency:** weekly automated run + daily human audit<br><br>**Proposed alert threshold :** ≥5 % relative drop in F1 *or* ≥10 pp rise in human-model discordance |
| **Shift**<br>Behaviour changes caused by *external* alterations to inputs or model artefacts | **Example:** Introduction of ambient documentation<br><br>**Methods:**<br>● **Template simulation tests** – Re-format gold notes into new EHR templates & Dictation-to-text “ambient” conversions<br>● **Cohort shift checks** (e.g., pre- vs. post-COVID or new ICI release)<br><br>**Frequency:** quarterly stress-test suite<br><br>**Proposed pass criterion:** F1 within ±3 pp of baseline on simulated set | **Examples:** New model release or tool access.<br><br>**Methods**<br>● **API version pinning**—re-run the full gold set whenever the vendor releases *any* new model ID<br>● **Shadow inference**: route 5 % traffic to additional model, compare outputs offline<br>● **Latency & cost tracking** during shadow runs<br><br>**Frequency:** monthly<br><br>**Proposed pass criterion:** all primary metrics within pre-set tolerance (+/-3 pp) *and* no increase in hallucination or unsafe content flags from human |

**Table 6: Governance Decision Matrix by Deployment Phase at the Project Level**

| Phase & Core Question | Key Decisions | Responsible (R) | Accountable (A) | Consulted (C) | Informed (I) |
|---|---|---|---|---|---|
| **Phase 0: Security & Legal**<br>*"Is it secure and legal?"* | Data access permissions & compliance | Security Officer | CMIO | Legal, Privacy | Clinical Teams |
| | PHI handling approach (prompts vs fine-tuning) | Privacy Officer | CIO | IRB, Legal, Clinical AI | All stakeholders |
| | Infrastructure security design | Cloud Ops | CIO | Security, Compliance | Clinical Teams |
| **Phase 1: Safety**<br>*"Is it safe?"* | Data types in scope | Clinical Informatics | Oncology PI | Compliance | Security |
| | Acceptable error thresholds | Clinical Informatics | Governance Board | Patient Safety | Clinical Teams |
| | Red-team testing protocol | Security | CMIO | Clinical Informatics | All stakeholders |
| **Phase 2: Efficacy**<br>*“Can it integrate into workflow?"* | Data flow architecture | Data Engineering | Clinical Informatics | Cloud Ops | Helpdesk |
| | UI/UX design for clinicians | Clinical Informatics | Oncology PI | End Users (CRC/Physicians) | IT Support |
| | Response time requirements | Data Engineering | Clinical Informatics | IT Operations | All Users |
| **Phase 3: Effectiveness**<br>*"Does it improve care?"* | Performance criteria for go-live | Clinical Informatics | Governance Board | Quality, Project Lead | Clinical Teams |
| | ROI metrics & tracking | Clinical Informatics | CMIO | Project Lead, End Users | Leadership |
| | Comparison to standard care | Clinical Informatics | Oncology PI | Clinical Teams | IRB |
| **Phase 4: Surveillance** *"Does it stay safe & effective?"* | Drift monitoring thresholds | ML Engineers | Clinical Informatics | Governance Board, Project Lead | Clinical Teams |
| | Model update approval process | Clinical Informatics | Governance Board | All stakeholders | All users |
| | Incident response protocol | Clinical Informatics | CMIO | Security, Legal | All stakeholders |

**Table 7:** Implications of Five Heavy Lifts for Clinical LLM Agent Deployment

| **Heavy lift** | **How model/algorithm builders typically think about this** | **Implications for project leaders** | **Implications for organisational leaders** | **Implications for academic researchers and model developers** |
|---|---|---|---|---|
| **Data integration** | Assume that a clean, unified dataset is readily available and that extraction, linkage, and governance are one-time pre-processing steps outside the core modelling effort. | Treat data integration as a primary design constraint: right-size latency to the use case (e.g., daily batches vs real-time feeds), work with institutional data teams and data concierges to refine cohorts in sandboxes, and ensure patient-level linkage and provenance before deployment. | Invest in centralised, governed research enclaves and templated “research zones,” with clear data ownership and access policies, so multiple projects can reuse secure, auditable EHR pipelines rather than building bespoke feeds. | Investigate how data architecture (e.g., enclaves, note chunking, embedding services), data quality, and traceability influence agent performance, safety, and clinician trust, rather than treating data as fixed input. |
| **Model validation and refinement** | Emphasise retrospective metrics and benchmark performance on static test sets, with limited prospective evaluation, red-teaming, or assessment of workflow fit. | Plan validation as an ongoing service: build staged roll-outs (retrospective gold standard, limited field use, silent pilot, then targeted deployment), involve end-users in human–computer interaction studies, and explicitly test hallucinations, evidence use, and other failure modes. | Provide secure sandboxes and centrally funded bridge engineers to harden prototypes, and institutionalise friends-and-family previews and red-teaming so agents are stress-tested before broader release. | Develop evaluation frameworks for agentic systems that extend beyond accuracy to include prospective agreement with clinicians, reliability across API updates, and impact on safety and workflow. |
| **Ensuring economic value** | Standard metrics and pipeline for economic value creation, focusing on per-call inference cost or assumed time savings rather than full lifecycle costs and institutional priorities. | Co-develop the economic case with the technical plan by separating fixed from variable costs, instrumenting dashboards for volume, cost, and latency, and framing value in terms of redeployed effort, safety, and research acceleration rather than simple labour substitution. | Centralise large, fixed infrastructure investments (e.g., cloud landing zones, data pipelines) while allocating variable usage costs to projects, and align AI investments with service-line and finance priorities. | Create standard approaches for measuring the total value of clinical LLM agents, including monitoring, refactoring, and human oversight—and for relating model and prompt choices to long-term cost and benefit. |

| **Managing model and data drift** | Apply drift concepts from static ML, focusing on occasional dataset shift and retraining, with limited attention to generative behaviour drift, vendor model changes, or evolving clinical documentation. | Design for continuous surveillance by pinning model versions, scheduling regular re-scoring against frozen gold sets, monitoring input distributions, and reviewing live outputs for emerging failure modes such as hallucinations or tone changes. | Stand up dedicated clinical AI governance or operations teams and shared monitoring infrastructure so drift detection, incident response, and remediation are coordinated across projects rather than handled ad hoc. | Evaluate complementary drift detection methods—embedding-based data drift, gold-set rescoring, and LLM-as-judge output audits—and identify thresholds and remediation strategies that are practical in clinical environments. |
|---|---|---|---|---|
| **Governance** | See governance mainly as a one-time approval step (IRB, security, compliance) and often treat privacy risks for LLMs as identical to those for traditional supervised models. | Integrate governance into the project charter through a RACI matrix, map decisions to lifecycle phases, and design prompts and workflows so PHI handling, logging, and escalation paths are explicit. | Maintain an enterprise AI governance board with multidisciplinary representation and lifecycle checkpoints, standardise documentation expectations, clarify PHI risk for training vs prompting, and plan for safe decommissioning and surveillance. | Empirically compare governance and oversight models, and refine guidance around privacy risk for clinical data, fine-tuned models, and prompt optimisation so evaluation and regulation keep pace with agentic systems. |

| **Supplementary Table 1: Demographics of Interview Participants** | | |
|---|---|---|
| | | |
| **Years of Experience** | **Frequency** | **Percentage** |
| 1-5 years | 3 | 14% |
| 6-10 years | 2 | 10% |
| 11-19 years | 5 | 24% |
| 20 or more years | 11 | 52% |
| **Primary Domain of Expertise** | | |
| Technical | 8 | 38% |
| Operational | 8 | 38% |
| Clinical | 5 | 24% |
| **Education Level** | | |
| Bachelors | 5 | 24% |
| Masters | 4 | 19% |
| MD or Doctorate | 12 | 57% |
| **Organizational Role** | | |
| Junior | 5 | 24% |

| Mid-senior | 4 | 19% |
|---|---|---|
| Director | 8 | 38% |
| Executive | 4 | 19% |